\documentclass[11pt,letterpaper]{article}
\usepackage{fuxi-report}

\renewcommand{\reportabstract}{
Large language model agents have shown promise in bioinformatics, but most existing systems focus primarily on producing final answers, treating planning,
tool use, and code execution as transient interactions. This design is poorly suited to long-horizon bioinformatics tasks, where conclusions must remain
connected to the data, computations, and intermediate evidence that support them. We introduce \textbf{Bioinfoysis}, a multi-agent harness that represents each request as a persistent, artifact-grounded analysis run. Bioinfoysis combines global planning with step-wise, evidence-driven replanning: the planner maintains an executable checklist and revises pending steps using structured handoffs returned after each worker execution. These handoffs bind intermediate
results to their responsible agent, checklist step, and plan generation, preventing stale evidence from being silently reused after replanning. A controlled runtime validates generated scripts, tables, and figures before they are used in downstream analysis or reporting, while role-specific context, persistent memory, and governed bioinformatics skills support reliable execution
over long analysis trajectories. We evaluate Bioinfoysis on BixBench and two question-answering tracks of LAB-Bench 2. On BixBench, Bioinfoysis achieves
state-of-the-art accuracy of 82.4\%. Across four underlying language models, Bioinfoysis increases average accuracy from 27.81\% to 64.13\% on SeqQA2 and
from 3.13\% to 31.25\% on DbQA2. These results demonstrate that reliable bioinformatics automation depends not only on model capability, but also on the
harness that governs planning, execution, memory, and evidence flow. We hope that the emergence of Bioinfoysis will play a driving and leading role in the development of the bioinformatics community. Our demo website can be seen in \url{https://report.bioinfoysis.com/}.

}

\begin{document}
\makereporttitle

\tableofcontents
\clearpage

\section{Introduction}

\begin{wrapfigure}{r}{0.5\textwidth}
    \centering
    \includegraphics[width=\linewidth]{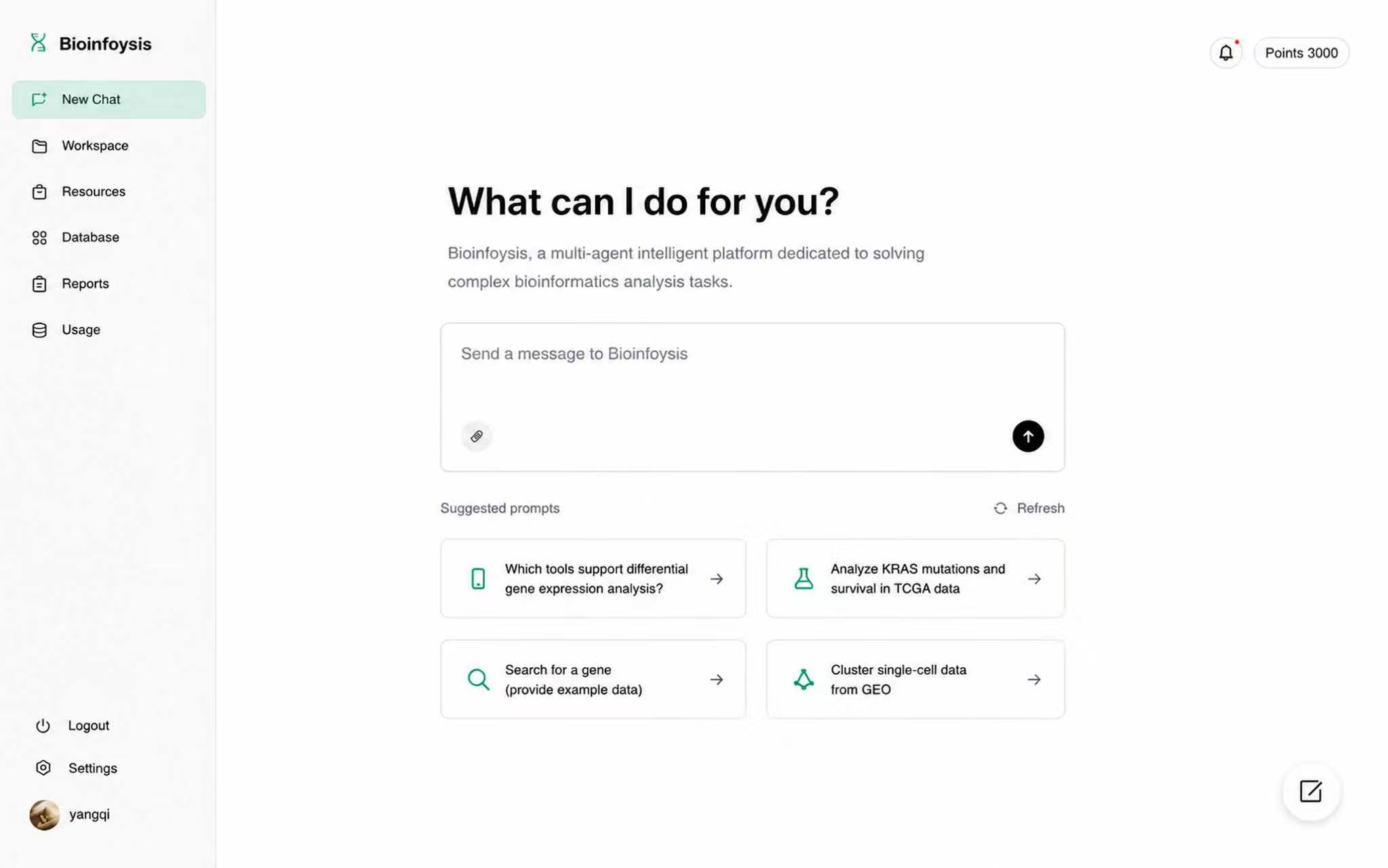}
    \caption{Overview of Bioinfoysis website.}
    \label{fig:overview}
\end{wrapfigure}
Bioinformatics has become an essential part of modern biological research \cite{huang2025biomni}. Many scientific questions can no longer be answered by inspecting experimental measurements directly. Researchers must transform raw data into evidence through computational analysis. As biological datasets continue to grow, this analytical process is becoming
longer and more difficult to manage. The main difficulty does not always lie in the absence of suitable algorithms. In many cases, the required statistical methods and software tools already exist. The difficulty lies in applying them correctly within a complete analysis.
Each decision depends on what happened before it. The interpretation of a column determines how the data are processed.
The preprocessing procedure determines which statistical model is appropriate. The result of that model determines what biological conclusion can be drawn. A small mistake at an early stage can therefore affect the entire analysis without making the final result appear obviously incorrect.

Large language models (LLMs) and agents have made this bottleneck more visible \cite{kong2026aiautoresearch, yang2025qwen3, singh2025openai, qiao2025signbot, zhang2025aixiv, langgraph2026overview}. On the one hand, they can write code, summarize biological concepts, call tools, and reason over the literature. On the other hand, bioinformatics workflows expose failure modes that are easy to miss in ordinary question answering: the model may select an incompatible package or generate a plausible report whose evidence cannot be traced back to the executed scripts. For scientific use, this distinction is critical. A final answer is useful only when the user can ask how it was produced, which data and tools were used, and whether the execution can be resumed or audited after an error. Bioinformatics, therefore, requires agents that behave less like chat interfaces and more like governed execution harness systems.

Recent biomedical agents have made substantial progress toward this goal. General biomedical agents integrate scientific reasoning, tool use, code execution, and access to databases \cite{weng2025dbms_ai} in broad research domains \cite{gao2024empowering,huang2025biomni}. More focused systems target the design of genetic perturbation experiments, interpretation of gene-sets, and multi-agent bioinformatics assistance
\cite{roohani2025biodiscoveryagent,wang2025geneagent,mehandru2025bioagents}. BioMedAgent further studies self-evolving tool-based workflows for biomedical data analysis \cite{bu2026biomedagent}. In parallel, benchmarks such as
BioCoder, BixBench, and BioAgent Bench have shifted evaluation away from static knowledge recall and toward executable code and multi-step analysis \cite{tang2024biocoder,mitchener2025bixbench,fa2026bioagentbench}. These works establish an important premise: bioinformatics agents should not merely answer questions; they should operate over data, tools, and methods. \textit{Long-horizon bioinformatics analysis, however, requires a system layer that connects these capabilities throughout the full task lifecycle.} The system must decide which agent should act next and preserve the state required by later steps. It must control where code is executed and keep generated files associated with the correct task. It must expose progress and failures during the analysis. It must also retain enough information to inspect, resume, and evaluate the analysis after the immediate conversation has ended. Agent frameworks, tool protocols, and multi-agent communication standards provide important building blocks for this layer \cite{anthropic2025multiagent,openai2025agentsdk,langgraph2026overview,anthropic2024mcp,google2025a2a}, but they do not by themselves define how a complete bioinformatics analysis should be organized.

We refer to this surrounding system as an \emph{bioinfomatics harness}. It is the execution layer that turns model outputs into a managed scientific process by coordinating planning, agent routing, runtime state, tool access, workspaces, intermediate results, artifacts, and reporting. The quality of the system is consequently determined not only by whether it returns the correct answer, but also by whether the path to that answer can be \textbf{executed, inspected, recovered, and verified.} This report introduces \textbf{Bioinfoysis}, a multi-agent bioinformatics harness for long-horizon bioinformatics analysis. Given a natural-language goal and a set of input files, Bioinfoysis creates a persistent run and converts the request into an executable checklist. Bioinfoysis is built around several capabilities that support this long-horizon process:

\begin{itemize}
\item \textbf{Coordinated multi-agent analysis.} Bioinfoysis organizes the analysis around a shared checklist and workflow state. Because these agents operate on the same analysis state, their outputs contribute to one continuous workflow rather than a sequence of independent conversations.

\item \textbf{Evidence-grounded and skill-guided bioinformatics execution.}
Bioinfoysis connects biological knowledge retrieval with executable bioinformatics procedures. Before computation, agents can search the web, query scientific resources, and use retrieval-augmented generation to obtain methodological guidance and biological evidence. During computation, basic tools provide operations such as inspecting files, inferring table schemas, checking sample overlap, searching databases, and executing code in a sandbox. Bioinformatics skills \cite{chen2025cedar} organize these operations into task-specific analytical procedures. 

\item \textbf{Role-specific context and persistent memory.}
Bioinfoysis constructs a focused context for each agent instead of repeatedly inserting the complete analysis history into the prompt. It combines recent run summaries, relevant artifacts, structured step results, and previously retained facts, and then presents different views to each agent based on their roles and current checklist steps. Long tool outputs are stored behind retrievable identifiers, allowing the prompt to use summaries while preserving access to the original content. After a run is completed, selected conclusions, evidence references, and artifacts are converted into structured long-term memory and can be retrieved for related tasks in later runs. 

\item \textbf{Structured handoffs and artifact continuity.}
After completing a checklist step, an agent produces a structured handoff containing its execution status, main conclusion, supporting evidence, references, generated artifacts, and process summary. Each handoff is associated with the responsible agent, the corresponding checklist step, and the version of the complete plan under which it was produced. Later agents can therefore continue from explicit results and files rather than reconstructing previous work from conversational text. When a new complete plan is generated, records from an earlier plan version are not returned as valid results for steps in the active plan.

\item \textbf{Trace-based audit and evaluation.}
Each Bioinfoysis run is linked to a structured trace, which records model and tool activity, execution status, latency, token usage, cost, retrieval summaries, and references to generated artifacts. A run-level audit record further connects the final answer to its input files, commands, skills, tools, outputs, and runtime environment. By combining these traces with benchmark scores and automatic evaluators, Bioinfoysis supports comparisons across models and harness configurations and helps localize failures to a particular stage of the analysis.

\end{itemize}

We evaluate Bioinfoysis on BixBench, a benchmark comprising real-world computational-biology scenarios that require agents to explore datasets, perform multi-step analyses, and interpret open-ended biological results \cite{mitchener2025bixbench}. Bioinfoysis achieves 82.4\% accuracy with Kimi-K2.6, 79.0\% accuracy with GPT-5.4-API and 77.0\% with DeepSeek-V4, placing it among the strongest bioinformatics agents in our comparison. We further evaluate the system on the SeqQA2 and DbQA2 tracks of LAB-Bench2. Across four model families, Bioinfoysis improves average accuracy from 41.65\% to 69.69\% on SeqQA2 and from 3.13\% to 27.60\% on DbQA2. These consistent gains across models and task types indicate that reliable bioinformatics performance is determined not only by the underlying LLM, but also by whether the surrounding harness can correctly interpret and route tasks, maintain context and memory, execute tools and skills, preserve intermediate artifacts, and validate results before producing an answer.

All in all, our results position Bioinfoysis as more than a tool-augmented language model: it provides a persistent and inspectable execution framework for long-horizon bioinformatics analysis. By connecting planning decisions, computational actions, intermediate evidence, and final conclusions within a unified run record, Bioinfoysis makes complex analyses easier to verify, recover, and extend. More broadly, this work suggests that progress toward reliable AI systems for computational biology will depend as much on the design of the agent harness and scientific execution infrastructure as on advances in foundation models themselves.

\section{Architecture}

\subsection{System Overview}

\begin{figure}[t]
    \centering
    \includegraphics[width=\linewidth]{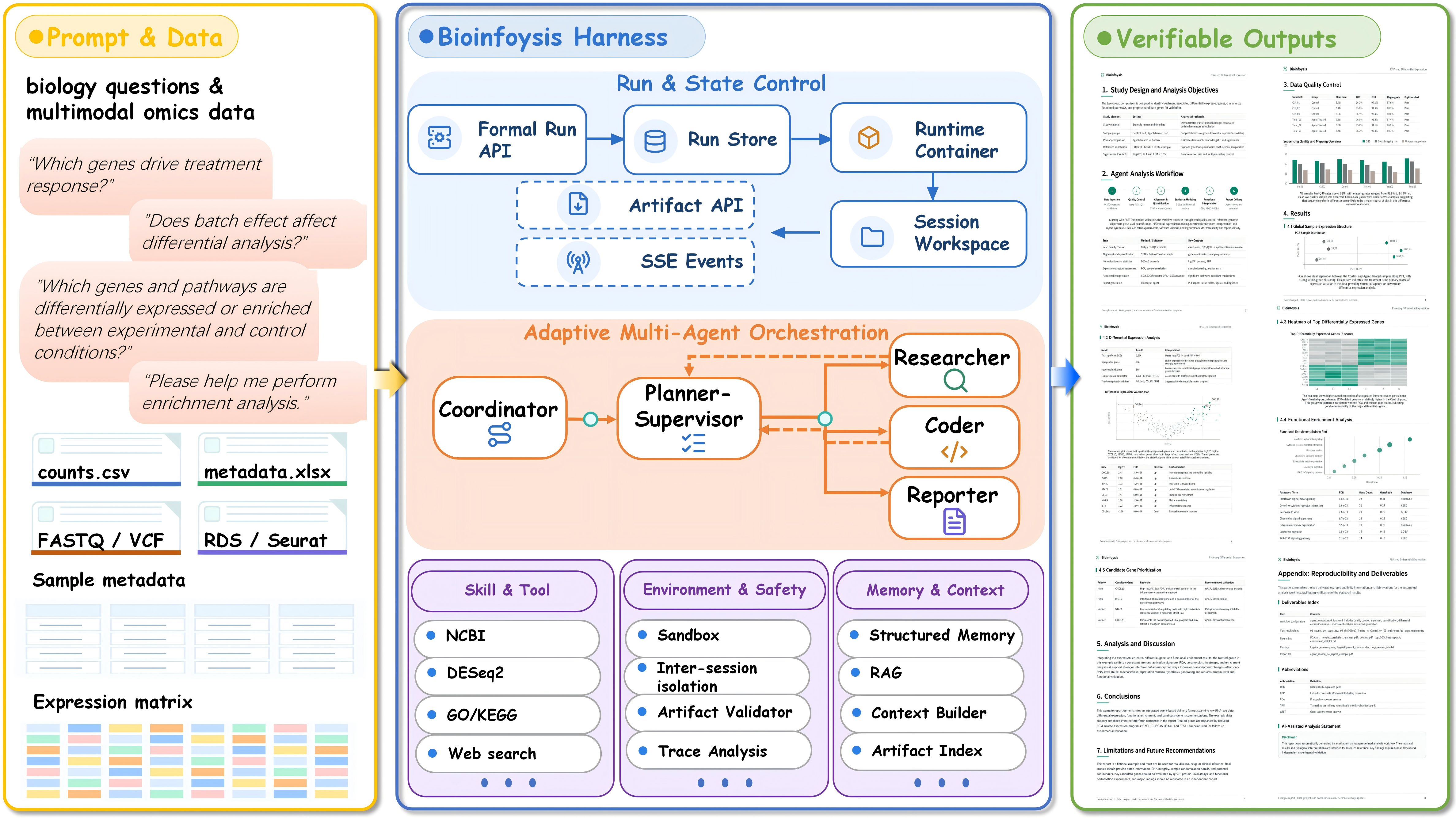}
    \caption{\textbf{Overview of the Bioinfoysis framework.} Bioinfoysis transforms bioinformatics questions and multimodal omics data into verifiable analysis reports through a persistent, adaptive multi-agent workflow.}
    \label{fig:bioinfoysis-framework}
\end{figure}

Bioinfoysis is a multi-agent harness for long-horizon bioinformatics analysis and scientific report generation. Figure \ref{fig:bioinfoysis-framework} presents the Bioinfoysis framework. Its purpose is not simply to connect a language model to a collection of tools. Instead, it governs how an open-ended request is converted into a plan, how that plan is executed against biological data and external evidence, and how the resulting computations are assembled into a report that can be checked against the underlying files. Each request is represented as a persistent run with explicit inputs, workflow state, intermediate results, and final deliverables. The primary system object is therefore the analysis process rather than the final chat message \cite{zhang2025aflow, zhao2026a2flow, qian2025scaling, li2026flow, zhang2025agenticcontextengineeringevolving, wang2026phantom}. The harness consists of three interacting layers: an \textit{Orchestration layer} for multi-agent planning and
checklist-based execution \cite{hong2024metagpt, dibia2024autogen, chen2024agentverse, agrawal2025gepa}; a \textit{Scientific Capability and Execution layer} for agent-specific tools, bioinformatics skills management, and isolated code execution, and artifact validation; and an \textit{Observability layer} for traces, analytics facts, and audit summaries. Together, these layers convert open-ended biological analysis requests into structured, inspectable, and reproducible scientific workflows.

The default execution path is organized around five functional roles:
\emph{coordinator}, \emph{planner}, \emph{researcher}, \emph{coder}, and
\emph{reporter}, together with an optional human-feedback gate \cite{yu2026habituation, qiao2026focus}.

\begin{itemize}
    \item The coordinator first determines whether the request requires bioinformatics analysis, resolves uploaded or previously generated files, and converts the user input into a compact, file-aware task description. 
    \item The planner first constructs an initial global checklist that specifies the main analytical objectives, worker assignments, dependencies, and expected outputs. This checklist serves as an executable scaffold rather than a frozen workflow. After the plan has been accepted, the same planner node enters \emph{planner--supervisor mode}, in which it uses newly produced evidence and execution feedback to refine the remaining steps before dispatching the next worker.
    \item The researcher provides complementary forms of scientific evidence. The researcher obtains biological and methodological information from the literature, NCBI resources, supported databases, and configured retrieval services. 
    \item The coder inspects the input data, checks schemas and sample relationships, writes and executes analysis scripts, and produces computational artifacts such as tables and figures in an isolated sandbox environment. 
    \item The reporter is retained as the final checklist step and is invoked only after the required evidence and artifacts have been produced. It synthesizes the validated step results into a final report while preserving references to their supporting sources and generated files. 
\end{itemize}
This separation of planning, evidence acquisition, computation, and reporting makes the workflow more controllable and inspectable than a single-agent analysis process.

\subsection{Framework}

\subsubsection{Orchestration Layer: Adaptive Planning and Execution}

The orchestration layer follows an adaptive plan-and-execute strategy. The
workflow begins with a coordinator that determines whether the request requires
bioinformatics analysis, resolves the relevant files, and prepares a compact,
file-aware task description. Based on this information, the planner constructs
an initial global checklist that specifies the main analytical objectives,
worker assignments, execution dependencies, and expected outputs. The checklist
provides a coherent execution scaffold, but it does not assume that every
analytical decision can be fixed before the data and intermediate results have
been examined.

After the checklist has been accepted, the same planner node enters
\emph{planner--supervisor mode}. Execution and planning then proceed
interleavedly. The planner--supervisor dispatches one checklist step to the
researcher or coder and receives a structured handoff containing the resulting
evidence, artifacts, uncertainty, and execution status. Before dispatching the
next step, it reasons over this newly produced information and the current
checklist. If the result reveals an unexpected data structure, insufficient
evidence, a recoverable failure, or an additional analytical requirement, it
may add a new step or refine the description of a future pending step \cite{yao2022react}.

This refinement is deliberately constrained. Completed and in-progress steps
are not rewritten, and the reporter remains the final checklist entry. The
workflow therefore combines the global consistency of an initial plan with the
flexibility of result-conditioned replanning. Rather than executing a fully
fixed pipeline, Bioinfoysis progressively develops the analysis as evidence is
produced while preserving the state and history of the accepted plan.

\subsubsection{Scientific Capability and Execution Layer: Tool- and
Skill-Guided Bioinformatics Analysis}

This layer provides the scientific capabilities required to transform a planned
step into biological evidence or a computed result. Instead of exposing the
same capabilities to every agent, Bioinfoysis assigns them according to the
agent's analytical responsibility. The researcher uses literature search, NCBI
resources, supported biological databases, document retrieval, and prior
evidence to collect the biological and methodological knowledge required by the
current step. The coder inspects the input data, identifies its structure and
sample relationships, selects an appropriate analysis procedure, executes the
computation, and produces result tables, figures, and scripts.

Tools provide the basic operations needed to search scientific resources,
inspect data, and execute analysis code. Skills organize these operations into
reusable bioinformatics procedures. A skill describes the assumptions, expected
inputs, recommended methods, software dependencies, execution guidance, and
expected outputs of a recurring analysis. For example, a differential-expression
skill \cite{wang2026safe} can guide the coder from count-matrix inspection and sample matching to
model fitting, statistical testing, and result visualization. Other skills
support procedures such as gene-set enrichment, single-cell differential
analysis, gene-identifier conversion, microbial variant analysis, methylation
analysis, and clinical statistics. Skills therefore contribute domain-specific
analysis guidance, rather than merely providing additional callable functions.

The resulting analysis is grounded in actual computation. Code is executed in a
controlled runtime associated with the current run, and generated scripts,
tables, figures, and logs are retained as analysis artifacts. Before a step is
accepted, Bioinfoysis checks that the reported artifacts exist and that
computation-dependent conclusions are supported by execution evidence or
quantitative results. When a recoverable problem occurs, such as an unexpected
table schema, missing column, sample mismatch, empty intermediate result, or
software dependency error, the system can return the failure information to the
coder for a bounded retry. In this way, tools determine what operations are
available, skills provide bioinformatics guidance on how to combine them, and
the execution environment ensures that the reported result is supported by an
actual analysis.

\subsubsection{Observability Layer: Run Evidence, Trace, and Evaluation}

The observability layer preserves the relationship between the final report and
the process that produced it. During a run, Bioinfoysis retains the selected
inputs, executed scripts, retrieved references, intermediate tables, generated
figures, step results, and execution failures. These records are organized by
run and linked to the checklist steps and agents that produced them. The final
report can therefore refer to concrete scientific evidence rather than relying
only on the language model's recollection of earlier tool use. The evidence and artifacts produced during each step are returned to the
planner--supervisor through a structured handoff. They therefore serve not only
as inputs to the final report, but also as feedback for refining the remaining
analysis.

Bioinfoysis also records the progression of the multi-agent workflow, including
agent execution, tool and skill use, artifact generation, model usage, failures,
and completion status. These records are projected into readable traces and a
structured audit summary that supports failure diagnosis, model comparison, and
benchmark evaluation. Large scientific files remain in the artifact workspace,
while the trace retains their identities and relationships to the corresponding
analysis steps.

This layer extends evaluation beyond final-answer accuracy. A benchmark result
can be examined together with the plan, intermediate evidence, executed
analysis, generated artifacts, token and cost information, and failure record.
Consequently, Bioinfoysis can distinguish errors caused by planning, retrieval,
data analysis, tool execution, or report synthesis. The same evidence structure
also makes it possible to compare different underlying language models within a
consistent analysis harness.

\section{Agent Infrastructure}
\label{sec:infrastructure}

The Agent infrastructure makes the bioinformatics workflow persistent. It stores the state of each analysis, prepares the context seen by each agent, executes code, preserves generated outputs, and records enough information to inspect or recover a run \cite{yuan2026sp, kuang2025learning, nie2026understanding}. These services are kept separate from the agent graph so that the same workflow can run on a single machine or a distributed deployment without changing its planning logic. Our infrastructure includes: A request enters through a run interface and is processed by the adaptive agent loop. Context, tools, skills, and the execution runtime support this loop. Artifacts and execution events are stored outside the prompts, while recovery and evaluation operate on the resulting run record. The following sections describe these components in detail.

\subsection{Persistent Runs and Controlled Execution}

Bioinfoysis stores each analysis as a persistent run. A run contains a stable identifier, the selected input files, its current workflow state, an ordered event history, and an index of generated artifacts. This representation allows a user or external platform to monitor an analysis, inspect a failure, and revisit its outputs after the live interaction has ended. The lifecycle distinguishes completed and failed runs from runs that are waiting for user input, cancellation, or a valid recovery point. Closing a live event stream therefore does not by itself mark the analysis as complete.

Each run executes in a controlled environment connected to a session workspace. The runtime uses a run-specific working directory and keeps the registered inputs separate from generated outputs. Before execution, it reconstructs the required inputs and checks their recorded identities. During execution, commands, scripts, and outputs remain associated with the active run. State-changing operations within the same session are serialized so that concurrent analyses cannot overwrite one another. When a valid workflow checkpoint exists, an interrupted run can continue under the same run identifier instead of being replaced by a new execution.

The same run interface is used in local and distributed deployments. A local deployment executes the workflow on one machine and stores objects on the filesystem. A distributed deployment separates request admission from execution: workers claim queued runs, maintain ownership while executing, and exchange inputs and outputs through shared storage. The deployment changes where computation takes place, but it does not change how the graph represents the run, the current checklist, the event sequence, or the generated artifacts.

\subsection{Role-Specific Context and Persistent Memory}

Bioinfoysis does not send the complete conversation and every previous tool result to every agent. At the beginning of a run, it builds a session context from recent conversation, summaries of completed runs, registered artifacts, stored memory \cite{weiretrieval, wang2026memguard, zhang2026memmark}, and recovery information. This context is then converted into separate views for the planner, researcher, coder, and reporter. The planner sees the task and current checklist, the researcher sees the current evidence need, the coder sees the active analysis step and relevant inputs, and the reporter sees the validated results and artifact references needed for synthesis.

Context within a run is centered on the active checklist step. Worker outputs are converted into structured handoffs that record the conclusion, supporting evidence, references, artifacts, uncertainty, and execution status. Compact evidence cards and reporter-facing result bundles are derived from these records. Before a previous result is reused, Bioinfoysis checks that it was produced by the expected agent for the expected checklist step and the current plan generation. Results from another step or an older complete plan are therefore not treated as valid evidence for the current analysis.

Large tool outputs are stored outside the prompt and replaced with a short preview and a stable result identifier. An agent can retrieve the complete content when needed, without carrying the full output through every later model call. When the planner context approaches its budget, Bioinfoysis progressively shortens low-value content, removes material that can be recovered elsewhere, and creates a structured summary only when necessary. Worker and reporter contexts mainly rely on role-specific selection and bounded evidence views. Active instructions, the current step, user clarifications, skill guidance, and newly retrieved evidence are protected.

Memory across runs is also restricted. Only completed runs can contribute evidence-backed facts to persistent session memory. Results from failed, cancelled, or waiting runs are not promoted as settled knowledge. Retrieval makes previous information available, while role-specific context construction determines which parts are relevant to the current agent.

\subsection{Artifact-Grounded Execution}

Generated outputs are managed as run artifacts rather than as paths mentioned only in an agent response. When the coder reports a script, table, file, or figure, Bioinfoysis resolves the path inside the permitted workspace and checks that the file exists. A computation-heavy step must also provide evidence that the analysis was executed or report quantitative results consistent with the requested output. Invalid paths and unsupported claims trigger a bounded correction instead of being attached to a successful step.

The workspace separates registered inputs, public scientific outputs, scripts, diagnostic logs, private tool results, memory, and trace data. Agents can access the internal materials required to continue the analysis, while users receive a restricted artifact view containing the intended deliverables. This separation prevents internal memory and raw execution payloads from being exposed as report outputs. It also allows the final report to refer to a stable run-scoped artifact instead of a temporary path on a particular worker.

Before the run is marked as complete, Bioinfoysis freezes and indexes its public artifacts. The index records each artifact's logical path, type, size, media type, and content hash. Distributed deployments can publish the corresponding file bodies to shared object storage. If a required table or figure cannot be published, the run is not reported as successfully completed. Optional diagnostic files can fail separately without invalidating the scientific outputs, and their failure records remain available for debugging.

\subsection{Trace, Audit, and Recovery}

Bioinfoysis records execution at several levels. The public event stream provides sanitized progress updates for users and external clients. The internal trace records agent calls, tool use, retrieval, artifact creation, failures, and workflow transitions. Service logs remain a separate source of low-level diagnostic information. Keeping these records separate prevents private context and raw tool outputs from leaking into the public progress stream while preserving the information needed to understand a failed run.

Internal events are converted into a readable run trace and structured statistics for agent nodes, tools, retrieval operations, artifacts, model usage, cost, and latency. When enabled, Langfuse displays these records under the same deterministic run identity. Bioinfoysis also creates an audit record that summarizes the request, input-file identities, executed commands, tools and skills, generated outputs, environment information, missing capture fields, and final response. Large datasets and figures remain in artifact storage; the trace and audit records retain links and metadata rather than copying their contents.

Recovery uses the same persistent state. If execution is interrupted, Bioinfoysis checks the durable run state, recent activity, current owner, session lock, and workflow checkpoint. The run resumes only when a valid checkpoint is available and continues under the same public run identifier. Publication and memory-update operations are protected from being applied twice after recovery. Cancellation follows the same principle: queued work can stop immediately, while running work stops at a safe workflow boundary \cite{jiang2026chaincaps} and may remain resumable. Recovery is therefore based on an explicit checkpoint, not simply on the disappearance of a worker process.

\section{Evaluation}
\subsection{Benchmark Suite}
We use BixBench \cite{mitchener2025bixbench} and LAB-Bench2 \cite{laurent2024labbenchmeasuringcapabilitieslanguage, laurent2026labbench2} as the core benchmark suite, as they evaluate complementary capabilities. BixBench assesses whether an agent can complete realistic, executable bioinformatics analyzes, while LAB-Bench2 assesses foundational competencies in biological research, including scientific knowledge, literature comprehension, protocol reasoning, sequence analysis, and database retrieval. Together, they measure both whether an agent can perform an analysis and whether it possesses the underlying capabilities required for biological research.

BixBench, introduced by FutureHouse and collaborators, is designed to evaluate whether bioinformatics agents can complete realistic, end-to-end analytical tasks. It comprises 61 analysis scenarios derived from scientific studies, yielding 205 open-ended questions across approximately 20 domains, including genomics, transcriptomics, differential expression, whole-genome sequencing, phylogenetics, sequence analysis, genomic variant analysis, epigenomics, proteomics, single-cell analysis, and integrative omics. Its basic unit is the \emph{capsule}, which organizes a biomedical hypothesis or analytical finding together with the corresponding input data, analysis code, conclusions, and gradable questions. The underlying workflows are implemented in Python, R, or Bash and are typically documented in Jupyter notebooks. Some capsules reproduce analyses from published studies, whereas others require agents to construct new analytical workflows from the provided data. Consequently, BixBench evaluates more than factual knowledge: an agent must inspect local datasets and project structures, understand or execute analysis code, perform statistical analyses, and report concrete outputs such as numerical estimates, genes, pathways, variant counts, sequence statistics, or genomic intervals.

LAB-Bench2, short for the Language Agent Biology Benchmark, is also introduced by FutureHouse to evaluate language models on foundational capabilities required for biological research. The benchmark contains 2,457 multiple-choice questions spanning eight categories and 30 subtasks. Its eight categories are LitQA, SuppQA, FigQA, TableQA, DbQA, ProtocolQA, SeqQA, and CloningScenarios. LitQA2 evaluates literature retrieval and the understanding of recent biomedical findings. For this report, we only focus on the DbQA and SeqQA.  DbQA tests the ability to query commonly used biological databases, while ProtocolQA evaluates protocol comprehension and error identification. SeqQA covers sequence-related reasoning tasks such as identifying open reading frames, designing PCR primers, predicting restriction fragments, and calculating GC content. 

\subsection{Main Results}

\subsubsection{Bixbench benchmark}

\begin{figure}[t]
    \centering
    \includegraphics[width=\linewidth]{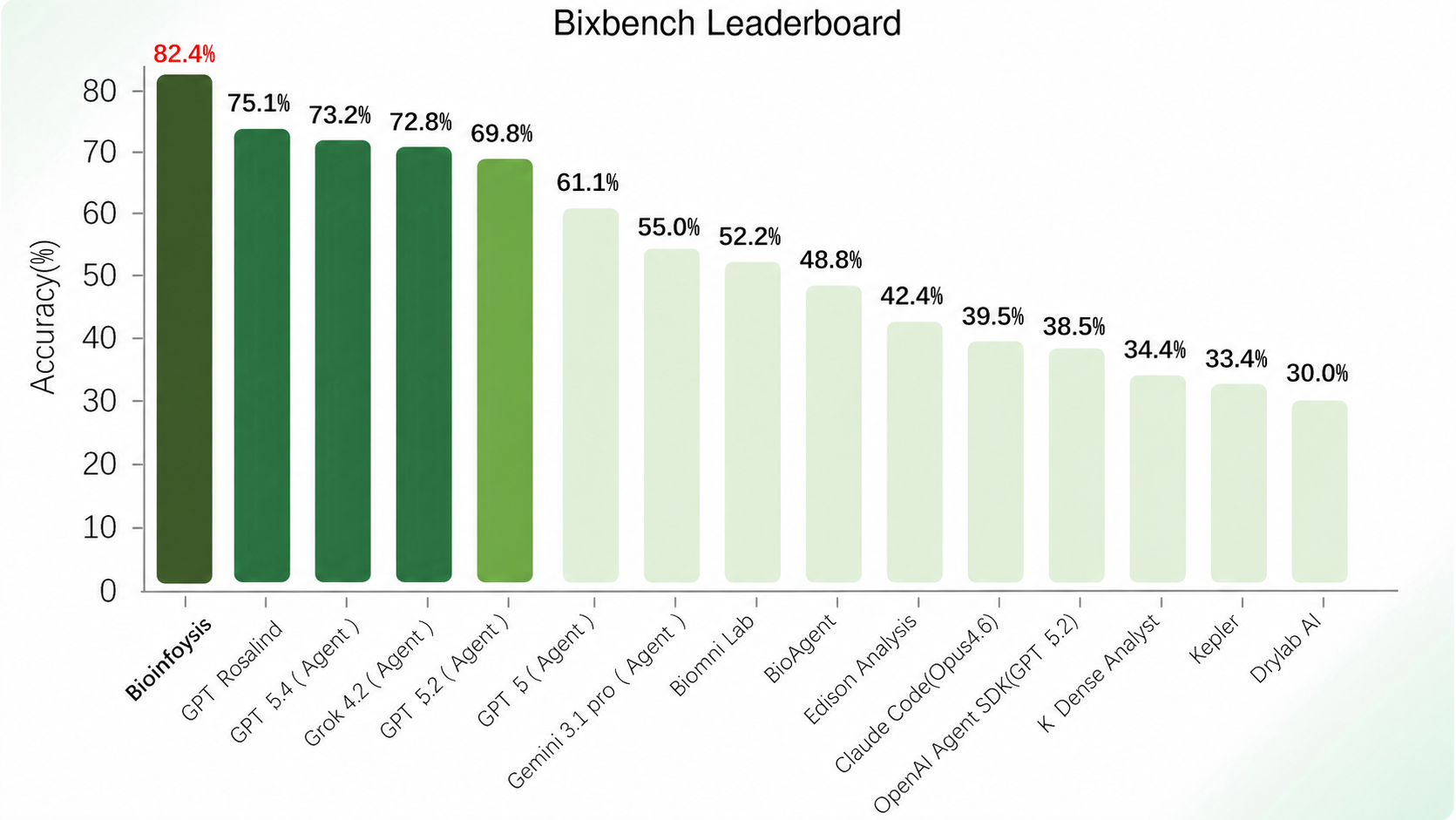}
    \caption{\textbf{BixBench leaderboard.}
Accuracy of Bioinfoysis and representative general-purpose,
tool-using, and bioinformatics-oriented agent systems on BixBench.
Bioinfoysis achieves the highest accuracy of 82.4\% (169/205),
outperforming the second-ranked system by 7.3 percentage points. Results
from systems with different models and execution configurations are
presented as system-level comparisons.}
    \label{fig:bixbench-leaderboard}
\end{figure}

\begin{table}[htbp]
\centering
\caption{Domain-wise evaluation results of Kimi-K2.6 on BixBench.}
\label{tab:bixbench-kimi-domain-summary}
\small
\begin{tabular}{lrrrr}
\toprule
Domain & Capsules & Correct & Questions & Accuracy \\
\midrule
Functional Genomics \& Pathway Enrichment
    & 9  & 22  & 31  & 70.97\% \\
Genomic Variant Analysis
    & 7  & 14  & 16  & 87.50\% \\
Transcriptomics \& Differential Expression
    & 10 & 30  & 42  & 71.43\% \\
Phylogenetics \& Comparative Genomics
    & 15 & 45  & 51  & 88.24\% \\
Epigenomics
    & 2  & 9   & 12  & 75.00\% \\
Clinical Research \& Biostatistics
    & 3  & 16  & 18  & 88.89\% \\
Bioimaging \& Quantitative Phenotyping
    & 4  & 21  & 21  & 100.00\% \\
Single-Cell Transcriptomics
    & 1  & 2   & 2   & 100.00\% \\
Proteomics
    & 1  & 4   & 4   & 100.00\% \\
Multi-Omics Integration \& Machine Learning
    & 1  & 2   & 2   & 100.00\% \\
Microbial Genomics \& Antimicrobial Resistance
    & 1  & 4   & 6   & 66.67\% \\
\midrule
\textbf{Overall}
    & \textbf{54}
    & \textbf{169}
    & \textbf{205}
    & \textbf{82.44\%} \\
\bottomrule
\end{tabular}
\end{table}

We compare Bioinfoysis with representative general-purpose language models, tool-using agents, and bioinformatics-oriented agent systems on the
original BixBench benchmark, which contains 205 questions derived from 61 real-world bioinformatics analysis capsules. Accuracy is defined as the
percentage of questions answered correctly according to the benchmark evaluation protocol. Bioinfoysis uses Kimi-K2.6 as its underlying agent model and enables the complete agent harness. Results for the comparison systems are collected from their reported BixBench
evaluations.

As shown in Figure~\ref{fig:bixbench-leaderboard}, Bioinfoysis achieves the highest accuracy of 82.4\%, correctly answering 169 of the 205
questions. It outperforms the second-ranked GPT Rosalind system by 7.3 percentage points and exceeds the baselines of GPT-5.4 Agent and Grok-4.2 Agent by 9.2 and 9.6 points, respectively \cite{liu2026lifescibench}. Compared with bioinformatics-oriented
systems, Bioinfoysis improves over Biomni Lab \cite{huang2025biomni} and BioAgent \cite{weidener2026rethinking} by 30.2 and 33.6 percentage points. The performance advantage is even greater than
that of earlier systems such as Edison Analysis, K-Dense Analyst \cite{li2025k}, Kepler, and Drylab AI, whose accuracies range from 30.0\% to 42.4\%. These results indicate that strong performance on BixBench cannot be attributed solely to the use of a capable underlying language model. Several agents based on general-purpose models, including the GPT-5.4 Agent, Grok-4.2 Agent, and the GPT-5.2 Agent, obtain precision between 69.8\% and 73.2\%, despite the use of competitive foundation models \cite{wang2026embedding}. In contrast,
Bioinfoysis combines its underlying model with an execution harness that preserves intermediate evidence, dynamically revises the analysis workflow, invokes task-specific bioinformatics procedures, and validates generated artifacts before producing the final answer, finally achieving SOTA performance.

As displayed in Table~\ref{tab:bixbench-kimi-domain-summary}, Bioinfoysis achieves an overall accuracy of 82.44\% (169/205), while its performance varies across biological domains. The system performs strongly on Bioimaging and Quantitative Phenotyping, correctly answering all 21 questions. It also achieves high accuracy in Phylogenetics and Comparative Genomics (88.24\%), Clinical Research and Biostatistics (88.89\%), and Genomic Variant Analysis (87.50\%). These results indicate that the framework can reliably execute workflows involving quantitative measurements, statistical
analysis, variant interpretation, and comparative analysis. The main errors are concentrated in Functional Genomics and Pathway Enrichment (70.97\%) and Transcriptomics and Differential Expression (71.43\%). Together, these two domains account for 21 of the system's 36 incorrect answers. Such tasks frequently require multiple dependent decisions, including sample
selection, statistical model specification, gene filtering, identifier conversion, and interpretation of enrichment results. Errors introduced at an
early stage can therefore propagate to the final answer. Bioinfoysis reaches 100\% accuracy in Single-Cell Transcriptomics, Proteomics, and Multi-Omics Integration and Machine Learning. However, these categories
contain only two to four questions each and should therefore not be interpreted
as evidence of uniformly solved domain capability. Generally, the results demonstrate broad cross-domain effectiveness.

Because differences in bioinformatics analysis conventions can substantially affect the resulting conclusions, our evaluation does not consider only whether the final answer produced for a single task matches the reference answer, but also considers the correctness of the final answer, the correctness of code execution, whether required intermediate and output files are successfully generated, and whether the resulting artifacts are sufficiently traceable to explain how the final answer was derived. Before using BixBench as a formal evaluation benchmark, we additionally conducted a benchmark audit. This was necessary because many BixBench tasks are not generated from a unified template, but instead consist of customized analytical questions originating from different capsules. Each capsule may adopt its own sample-filtering rules, denominator definitions, contrast directions, model formulas, database versions, preprocessing procedures embedded in notebooks, and answer formats. Consequently, a discrepancy between an agent's answer and the BixBench ideal answer does not necessarily indicate that the agent's analysis is incorrect. The question itself may admit multiple valid interpretations, or the official answer may depend on implicit notebook logic that is not specified in the task description. For details, please refer to the Appendix \ref{lab: bixbench_extended_evaluation}.

\subsubsection{Labbench2 benchmark}

\begin{table}[t]
    \centering
    \caption{\textbf{LABBench2 success rates on SeqQA2 and DbQA2.}
    Bioinfoysis is compared with the corresponding base-model setting.
    SeqQA2 contains 400 questions and DbQA2 contains 86 questions.
    Average denotes the macro-average accuracy across the four model columns.}
    \label{tab:labbench2_seqqa_dbqa}
    \resizebox{\linewidth}{!}{%
    \begin{tabular}{llccccc}
        \toprule
        Benchmark & System & Kimi & GPT-5.5 & DeepSeek & GLM & Average \\
        \midrule
        \multirow{2}{*}{SeqQA2}
        & Bioinfoysis
        & \textbf{66.00\%}
        & \textbf{75.00\%}
        & \textbf{68.50\%}
        & \textbf{69.25\%}
        & \textbf{69.69\%} \\
        & Base model
        & 35.50\%
        & 64.75\%
        & 31.61\%
        & 34.75\%
        & 41.65\% \\
        \midrule
        \multirow{2}{*}{DbQA2}
        & Bioinfoysis
        & \textbf{30.20\%}
        & \textbf{37.20\%}
        & \textbf{22.10\%}
        & \textbf{20.90\%}
        & \textbf{27.60\%} \\
        & Base model
        & 1.60\%
        & 7.80\%
        & 3.10\%
        & 0.00\%
        & 3.13\% \\
        \bottomrule
    \end{tabular}%
    }
    \vspace{2pt}

    \begin{minipage}{0.98\linewidth}
        \footnotesize
        SeqQA2 uses Kimi-K2.6, GPT-5.5, DeepSeek-V4-Pro, and GLM-5.2.
        For DbQA2, the reported base models are Kimi-K2.6, GPT-5.5,
        DeepSeek-V4-Flash, and GLM-5.2, respectively. Therefore, the
        DeepSeek result on DbQA2 is a system-level comparison rather than
        a strictly model-matched comparison.
    \end{minipage}
\end{table}

\begin{figure}[t]
    \centering
    \includegraphics[
        width=\textwidth,
        trim={0 0 0 0},
        clip
    ]{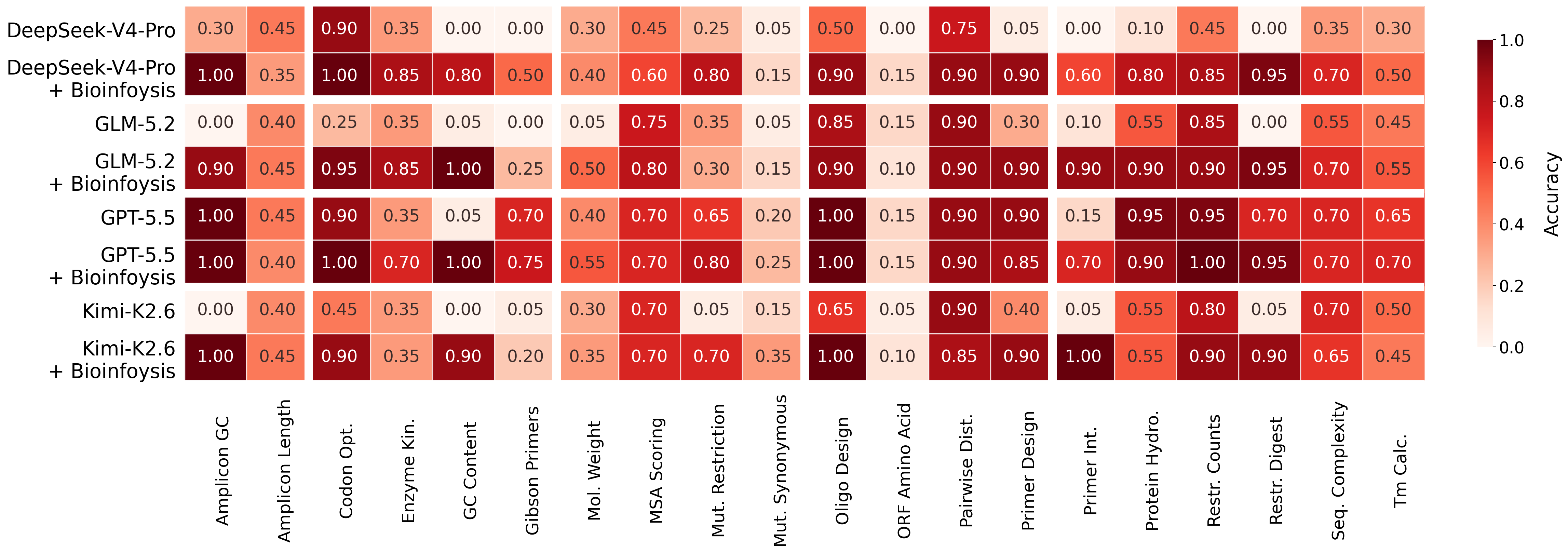}
    \caption{\textbf{Category-level performance on SeqQA2.} Each cell reports the success rate for one of the 20 SeqQA2 task categories under a particular model and system configuration. Darker red indicates higher accuracy. For each underlying model,
    the base-model result is shown alongside the corresponding Bioinfoysis result.}
    \label{fig:seqqa2_heatmap}
\end{figure}

We further evaluate Bioinfoysis on two question-answering tracks of
LAB-Bench 2: SeqQA2 and DbQA2. Four language models are evaluated under
two settings: direct inference with the base model and agent-based
analysis with the Bioinfoysis harness. Accuracy is defined as the
percentage of questions answered correctly. Table~\ref{tab:labbench2_seqqa_dbqa}
summarizes the overall results, while
Figure~\ref{fig:seqqa2_heatmap} presents the category-level performance
on SeqQA2.

On SeqQA2, Bioinfoysis consistently improves all four evaluated models,
raising their accuracies to between 66.00\% and 75.00\%. In comparison,
the corresponding base-model accuracies range from 31.61\% to 64.75\%.
The improvement is particularly substantial for DeepSeek, GLM, and Kimi,
which gain 36.89, 34.50, and 30.50 percentage points, respectively.
GPT-5.5 starts from a considerably stronger base-model accuracy of
64.75\% and further improves to 75.00\%, achieving the best overall
performance with Bioinfoysis. Its smaller gain of 10.25 percentage
points indicates that a strong base model can already solve a substantial
fraction of SeqQA2 through direct reasoning, while the structured workflow
still provides complementary benefits. The large improvements for Kimi, DeepSeek, and GLM suggest that
Bioinfoysis can effectively leverage the advantages of its harness to provide base models with extensive biological knowledge and structured problem-solving strategies. Many SeqQA2 questions can be converted into deterministic
operations, and Bioinfoysis decomposes these questions into explicit steps, executes the required operations programmatically, and preserves intermediate results for verification. Despite the consistent gains, the agent accuracies remain below 80\%,
showing that executable analysis alone is insufficient for fully solving
SeqQA2. Some failures arise when the agent constructs a scientifically
plausible workflow that does not conform to the assumptions or exact
criteria encoded by the official validator. Other errors result from
misinterpreting the requested calculation scope or biological definition. Answers can also be rejected because of incorrect units, numerical
precision, sequence formatting, or output representation, even when the
main analytical procedure is largely correct. 

The category-level heatmap in Figure~\ref{fig:seqqa2_heatmap}
shows that the benefits of Bioinfoysis extend across a broad range of
sequence-analysis tasks rather than being concentrated in only a few
categories. The most pronounced improvements are observed in restriction
digest analysis, GC-content analysis, amplicon-GC design, primer-interaction
analysis, and primer design. These tasks benefit from translating biological
requirements into explicit sequence operations and validating intermediate
computational results. By contrast, open-reading-frame amino-acid analysis,
synonymous-mutation reasoning, and Gibson-primer design remain among the most
challenging categories, reflecting the difficulty of multi-stage sequence
transformation and constraint-sensitive sequence design. Importantly, our manual audit shows that several design-oriented SeqQA2 questions admit multiple biologically valid answers. In such cases, an answer may satisfy the stated biological and sequence constraints while differing from the reference answer or the assumptions encoded by the official validator. We therefore supplement the automatic evaluation with manual inspection and constraint-based re-evaluation. The audit protocol and revised evaluation criteria are detailed in Appendix~\ref{lab: Labbench2_extended_evaluation}.

DbQA2 remains substantially more challenging for direct model inference than for agent-based analysis. The four base models achieve an average accuracy of only 3.13\%, whereas Bioinfoysis increases the average accuracy to 27.60\%, representing an absolute improvement of 24.48 percentage points. GPT-5.5 achieves the highest accuracy with Bioinfoysis at 37.20\%, followed by Kimi-K2.6 at 30.20\%, DeepSeek-V4-Flash at 22.10\%, and GLM-5.2 at 20.90\%. The higher accuracies primarily arise when the agent successfully completes the full workflow: identifying the appropriate data source, retrieving the relevant database records, and inferring the final answer. Models with stronger retrieval capabilities and more reliable workflow completion therefore achieve better overall performance. The remaining failures mainly stem from three sources. First, infrastructure and execution issues, including timeouts when downloading or processing large datasets, sandbox creation failures, and incomplete code execution, prevent some tasks from reaching the answer stage. Second, retrieval failures occur when specialized or version-specific resources cannot be accessed reliably, causing the agent to abstain under its conservative failure policy. Third, even when the required data are retrieved, errors may arise from selecting the wrong field or record, applying an incorrect denominator or analytical definition, or performing an inaccurate calculation. These results indicate that performance on DbQA2 depends not only on the underlying model’s reasoning ability \cite{wei2026think}, but also on the reliability of the end-to-end retrieval, execution, and validation pipeline.

\subsection{Ablations Study}
\label{sec:ablation}

\paragraph{Experimental setup.}
We evaluate the contribution of four components of Bioinfoysis on all 205 BixBench questions: different LLM model, ReAct-style planning, persistent memory, and governed bioinformatics skills. In \emph{Plan-Execute}, the initial plan is fixed before execution and cannot be revised in response to intermediate results. In \emph{w/o memory}, Bioinfoysis does not include an agent memory module and relies solely on the current context. In \emph{w/o bioinfomatics skills}, the skills include only those related to basic bioinformatics. In \emph{w/o all skills}, the skill catalog is removed entirely (zero skills loaded). We report both the evaluator's average score on a binary accuracy metric, where a question is counted as correct according to the benchmark evaluator. All configurations are evaluated on the same 205 questions.

\begin{table}[t]
    \centering
    \caption{Performance of Bioinfoysis on BixBench across different agent models and component ablations. The first three rows evaluate the harness with different agent models, while the remaining rows ablate ReAct-style planning, memory, and basic bioinformatics skills using GPT-5.4. $\Delta$ Acc. denotes the absolute accuracy change, in percentage points, relative to the full GPT-5.4 configuration.}
    \label{tab:bixbench_ablation}
    \resizebox{\linewidth}{!}{%
    \begin{tabular}{l l r r r r}
        \toprule
        Configuration & Agent model & Skills loaded & Accuracy (\%) &
        Correct & $\Delta$ Acc. \\
        \midrule
        Bioinfoysis & Kimi-K2.6 & 78 & \textbf{82.44} & \textbf{169/205} & \textbf{$+3.41$} \\
        Bioinfoysis & GPT-5.4 & 78 & 79.02 & 162/205 & $0.00$ \\
        Bioinfoysis & DeepSeek-V4-Pro & 78 & 77.07 & 158/205 & $-1.95$ \\
        \quad w/o ReAct-style planning
            & GPT-5.4 & 78 & 72.20 & 148/205 & $-6.83$ \\
        \quad w/o memory
            & GPT-5.4 & 78 & 56.59 & 116/205 & $-22.44$ \\
        \quad w/o bioinformatics skills
            & GPT-5.4 & 30 & 60.00 & 123/205 & $-19.02$ \\
        \quad w/o skills
            & GPT-5.4 & 0 & 42.44 & 87/205 & $-36.59$ \\
        \bottomrule
    \end{tabular}%
    }
\end{table}

We aim to investigate two questions: (1) \textit{Which components affect the performance of Bioinfoysis?} (2) \textit{Does the proposed agent harness deliver consistently strong performance across different models?} We observe several experimental findings from Table \ref{tab:bixbench_ablation}. 

(1) \textbf{Effect of ReAct-style planning.} Replacing ReAct-style planning with a Plan-Execute mode reduces accuracy from 79.02\% to 72.20\%, an absolute drop of 6.83 percentage points, and decreases the number of correct answers from 162 to 148. This result suggests that the ability to revise pending steps after observing intermediate evidence is useful for long-horizon analyses, where an initial plan may become invalid after file inspection, tool execution, or partial result validation. 

(2) \textbf{Effect of persistent memory.} Disabling memory produces the largest decline among the model-matched variants.
Accuracy decreases from 79.02\% to 56.59\%, corresponding to 22.44 percentage points and 46 fewer correct answers. In the harness, persistent memory records the selected inputs, completed operations, quantitative results, and paths to generated artifacts in structured step results. Later agents can therefore retrieve the exact result rather than infer it again from the remaining conversation. Without persistent
memory, these details exist only in the transient context. After context compression or an agent handoff, a later agent may know that an analysis was
completed but lose the exact sample filter, output table, column name, or numerical result. It may consequently repeat the analysis with different
settings, inspect the wrong artifact, or report an approximate value instead of the computed answer. The large and strongly asymmetric paired degradation
therefore indicates that persistent memory is particularly important for maintaining consistency across multi-step analysis and final-answer synthesis. 

(3) \textbf{Effect of governed skills.}
The skill ablations reveal that both domain-specific and general-purpose skills contribute substantially to performance. Removing the 48 bioinformatics-specific skills while retaining the remaining 30 skills reduces accuracy from 79.02\% to
60.00\%. This result indicates that general tool access alone is insufficient for many BixBench tasks. Bioinformatics skills provide task-specific procedures for operations such as selecting suitable analytical methods, configuring
domain tools, checking intermediate outputs, and extracting the quantities requested by the benchmark. Without this guidance, the agents must reconstruct these procedures from the task description and the language model's internal
knowledge, increasing the likelihood of inappropriate method selection, incorrect parameters, or incomplete result validation. Removing the remaining 30 skills further decreases accuracy from 60.00\% to 42.44\%. The remaining skills therefore provide capabilities that are complementary to specialized bioinformatics knowledge, such as structured file inspection, execution management, artifact handling, and result verification. These operations are particularly important in BixBench because obtaining a correct answer often requires not only knowing an appropriate biological method, but also locating
the correct input, executing the workflow successfully, and extracting the requested value from its outputs. 

(4) \textbf{Performance across agent models.} With the complete harness and the same set of 78 loaded skills, all three agent
models achieve accuracies above 77\%. Kimi-K2.6 obtains the highest accuracy at 82.44\% (169/205), followed by GPT-5.4 at 79.02\% (162/205) and
DeepSeek-V4-Pro at 77.07\% (158/205). The difference between the best and worst models is 5.37 percentage points. These results suggest that Bioinfoysis is not tightly coupled to a single
language model. Its checklist-based orchestration, structured handoffs, persistent memory, and governed skills provide a shared execution structure
within which models from different families can complete the benchmark. Nevertheless, the remaining variation shows that the underlying model still
affects planning, tool selection, interpretation of intermediate results, and final-answer synthesis. Kimi-K2.6 performs 3.41 percentage points above GPT-5.4, whereas DeepSeek-V4-Pro performs 1.95 points below it. Thus, the harness narrows the dependence on model choice but does not eliminate it.

The ablation results answer the two motivating questions. First, all four evaluated components make measurable contributions to Bioinfoysis. Persistent memory has the largest effect among the individual model-matched component ablations, with a 22.44-point reduction when removed. Eliminating adaptive ReAct-style planning causes a smaller but still substantial reduction of 6.83 points. The skill experiments display that the complete governed skill layer is equally central to the system: removing bioinformatics-specific skills
reduces accuracy by 19.02 points, while removing all skills reduces it by 36.59 points. In addition, the harness maintains strong performance across three different agent models, achieving 77.07\%--82.44\% accuracy with a range of only 5.37 percentage points. This cross-model consistency indicates that the framework's effectiveness is not attributable solely to one particular model.
Instead, performance arises from the interaction between the model and the surrounding harness: adaptive planning revises the workflow in response to intermediate observations, persistent memory preserves exact evidence across agent handoffs, and governed skills translate domain procedures into reliable execution paths. The model remains important, but the larger component-ablation
gaps show that system-level design is a major determinant of performance on long-horizon bioinformatics tasks.

\section{Conclusion, Limitations, and Future Work}
\subsection{Limitations} 

\paragraph{\bf Limited Tool and Skill Coverage.} Bioinfoysis’s analytical capabilities are constrained by the range of integrated tools and the coverage of its skill modules. For relatively standardized tasks, such as differential expression analysis, functional enrichment, and routine quality control of omics data, the system can organize workflows according to established analytical procedures. However, the current skill modules may not adequately support custom statistical models, specialized sequencing platforms, rare species, non-standard reference genomes, emerging algorithms, or highly customized analytical requirements. Even when a task is technically executable, the default method may not be optimal for the specific research question.

\paragraph{\bf Limited Evaluation Scope.} Although Bioinfoysis improves the structure, traceability, and recoverability of long-horizon bioinformatics analyses, several limitations remain. First, our empirical evaluation currently focuses primarily on BixBench and LAB-Bench2. Although these benchmarks cover a diverse range of bioinformatics workflows, performance on a limited set of benchmarks cannot fully characterize the reliability of the system in different data modalities, computational environments, and open-ended research settings. In addition, the benchmark primarily assesses the accuracy of final answers and does not capture all dimensions of scientific quality, such as the suitability of the statistical model, the robustness of the biological interpretation, or the reproducibility of results across alternative implementations.

\paragraph{\bf Lack of Autonomous Scientific Discovery.} Bioinfoysis currently does not provide comprehensive autonomous research. Its primary objective is to plan and complete bioinformatics tasks according to research questions, data, and analytical requirements specified by the user, rather than to autonomously formulate scientific questions, generate testable hypotheses, and conduct continuous open-ended exploration. Although Bioinfoysis can adapt subsequent analytical steps based on intermediate results, such adaptation remains within the task scope defined by. It should not be regarded as autonomous determination of research directions. For example, when an analysis reveals anomalous patterns, potential novel biomarkers, or observations inconsistent with the initial hypothesis, the system typically does not independently formulate competing hypotheses, design new validation experiments, or accumulate evidence across multiple analytical cycles.

\subsection{Future Work}
We will first expand the breadth and reliability of Bioinfoysis's
scientific capabilities. We plan to extend the current skill library to cover
more specialized data modalities, sequencing platforms, organisms, reference
genomes, and emerging analytical methods. Rather than simply integrating more
tools, each skill will explicitly define its applicable inputs, methodological
assumptions, configurable parameters, expected artifacts, and known failure
modes. We will also support alternative analytical strategies for the same
task, allowing users to compare methods and select workflows that better match
their specific research questions. In addition, we will expand the evaluation of Bioinfoysis beyond BixBench and LAB-Bench2. Future evaluations will include end-to-end research projects with
raw data, incomplete metadata, heterogeneous computational environments, and multiple scientifically reasonable analysis paths. In addition to final-answer accuracy, we will assess methodological appropriateness, biological validity,
artifact reproducibility, computational cost, and failure recovery. Finally, we aim to evolve Bioinfoysis from a user-directed analytical agent into an AI scientist for autonomous bioinformatics research \cite{tang2026ai, yuan2025dolphin, agarwal2026autodiscovery, xia2026sr, yamada2025ai}. Future versions will support the full research lifecycle: identifying open scientific questions, formulating testable hypotheses, designing analysis plans, retrieving relevant literature and datasets, executing computational experiments, interpreting results, and iteratively refining hypotheses based on accumulated evidence. We hope Bioinfoysis can conduct multi-cycle research in which each round of evidence motivates unexplored questions and experiments.

\subsection{Conclusion}

We present Bioinfoysis, a multi-agent framework for long-horizon bioinformatics analysis that integrates adaptive planning, persistent memory, governed skills, controlled execution, and artifact-grounded reporting. On BixBench, Bioinfoysis achieves $82.44\%$ accuracy and maintains strong performance across different underlying models. On the SeqQA and DbQA of LAB-Bench2, Bioinfoysis also achieves substantial improvements in task completion rates compared to the popular base models. Ablation results further demonstrate the importance of persistent memory, bioinformatics skills, and ReAct-style planning. Overall, these results illustrate that reliable bioinformatics agents depend not only on capable language models, but also on an execution harness that preserves evidence, coordinates specialized agents, and produces verifiable scientific outputs.

\newpage
\bibliographystyle{IEEEtran} 

\bibliography{reference}

\newpage
\appendix
\section{Author List and Acknowledgments}

\textbf{Core Contributors:}
Qingyang Shao,
Xin Zhang,
Zhouyang Yuan,
Xianying Chen.

\textbf{Contributors:}
Yujia Xiang,
Zihao Yang,
Tong Ye,
Yangqi Zhang,
Jiakang Xu,
Xiaoqing Yan,
Xuan Luo,
Keyi Li,
Enci Fan,
Kai Kang,
Zhuohan Liu,
Xingyu Jin,
Chunran Teng,
Tao Li,
Xinyu Lyu$^*$,
Minghui Wang$^*$,
Wenfeng Li$^*$,
Yidan Gao$^*$,
Siyu Liu,
Mingrui Luo,
Zhu Liang.

\textbf{Project Leader:}
Guanren Qiao (guanrenqiao1@link.cuhk.edu.cn),
Zhiping Xu (zhiping\_xu@dascience.cn).

$^*$ indicates individuals who have since left the team.
\section{Extended Evaluation Details}

\subsection{Extended Evaluation on Bixbench}
\label{lab: bixbench_extended_evaluation}

We manually audited both the BixBench summary table and the local capsule archive. The summary table contains 205 parsed records, and every record can be matched to a corresponding local capsule ZIP archive through the \texttt{data\_folder} field in the original BixBench JSON. Our manual annotations indicate that BixBench cannot be cleanly divided into only `correct' and `incorrect' questions; it also contains questions whose answers depend on interpretation, admissible answer ranges, or benchmark-quality issues. We therefore do not include every question directly in a strict exact-match evaluation. Instead, we distinguish among questions suitable for strict scoring, questions requiring multi-answer or range-based scoring, and questions requiring benchmark-quality analysis \cite{sun2026accuracymeasuringbiasacknowledgment}.

To scale this auditing procedure, we built a \textbf{Prompt Chain} on top of Bioinfoysis. The workflow consists of three steps. Step A opens the capsule, inspects the files, and extracts executable evidence. Step B performs an independent verification by comparing the reproduced results against the original BixBench fields and the literal requirements of the question, rather than simply trusting the official notebook. Step C produces a structured judgment containing fields such as \texttt{is\_error}, \texttt{answer\_verdict}, \texttt{reasoning\_verdict}, \texttt{error\_type}, \texttt{correct\_answer}, \texttt{evidence}, and \texttt{confidence}.

This design is important because reproducing the stored ideal answer with the official notebook does not necessarily imply that the ideal answer correctly addresses the question itself. The notebook may itself encode an incorrect sample-selection rule, analytical convention, or interpretation. Under this setting, Bioinfoysis is used not only as a task-solving agent, but also as a benchmark-auditing agent. Given a capsule archive, Bioinfoysis can inspect the file structure, execute scripts, reproduce intermediate tables, and generate artifacts that either support or challenge the official answer. The output of the prompt chain is therefore not a single-turn conversational judgment, but an evidence-backed audit record with a traceable chain of evidence. This also introduces a second dimension for evaluating Bioinfoysis: the system should not only solve well-defined benchmark tasks, but also identify defects in the benchmark itself and provide reproducible evidence supporting those findings.

Across the 15 analysis domains and 205 BixBench tasks reviewed here, disagreements arose from five distinct sources: underspecified task requirements, mismatches between the written question and the reference notebook, changes in software or external databases, defects in packaged data or provenance, and ordinary agent errors. These sources should not be treated as equivalent. In some tasks, several workflows are scientifically defensible; in others, the reference implementation is inconsistent with the question or data; and in comparatively stable tasks such as the ENO1 proteomics queries, disagreement is more likely to reflect an implementation or interpretation error.

The audit revealed several recurring issues, including ambiguous denominator definitions, hidden sample-filtering criteria, unclear set definitions such as whether an intersection or union is intended, database-version-dependent statistical results, species or cohort confusion, misinterpretation of result-table fields, and cases in which the official notebook reproduces the stored ideal answer but does not actually answer the literal question being asked. For example, some CHIP variant questions depend on whether intronic, intergenic, or UTR variants are excluded; some differential-expression questions depend on whether ``across all comparisons'' refers to an intersection or a union; and in some enrichment-analysis tasks, the BixBench ideal answer is not the term with the smallest adjusted (p)-value obtained when the analysis is rerun.

\begin{itemize}

    \item PhyKIT outputs may vary with the input alignment, summary statistic, group order, or selected output field. Some outputs depend on which PhyKIT field is read, which alignment is used, whether MAFFT or ClipKIT alignments feed downstream metrics, how gaps are summarized, whether orthologs are paired before comparison, and whether all available trees or a hidden subset is used.
    \item DESeq2 and enrichment tasks are the largest source of mixed errors. The reviewed rows show that results change with the selected samples, model design, DEG threshold, raw-(p) versus adjusted-(p) filtering, LFC shrinkage, gene-symbol mapping, background universe, enrichment library, and software or database version. Several official answers are reproducible only under a particular hidden or old environment. In ASXL1 and several GO, KEGG, and Reactome rows, different versions of clusterProfiler, GSEApy, Enrichr, KEGG, Reactome, and PyDESeq2 change adjusted (p)-values, odds ratios, ranked pathways, or DEG counts.
    \item Clinical regression estimates are sensitive to the case definition, the handling of repeated records from the same participant, the analysis population, category coding, and reference groups. In the BCG analysis, the question asks about COVID-19 severity, but the reference notebook defines each participant's severity as the highest value across all adverse-event records, including injection-site events unrelated to COVID-19. Restricting the outcome to COVID-19-related episodes measures a different clinical endpoint and can change both the size and direction of the estimated association. The stated outcome therefore does not match the case definition used in the reference notebook, and it remains unclear which event terms and repeated records should be included.
    \item For CHIP variants, the result changes depending on whether the analysis counts table rows, sample-specific calls, unique genomic variants, or affected patients, as well as on the filtering and deduplication rules. In the supplied CHIP table, the same variant can appear in several rows for the same sample, often because it has more than one annotation. Counting rows treats each entry as a separate variant, whereas deduplicating by sample, genomic position, and allele counts that variant only once within each sample. Because the question does not define the counting unit, the term ``variants'' may refer either to spreadsheet records or to biological variant calls.
    \item In colony morphology analyses, images, technical repeats, and independent experimental replicates may each be treated as the statistical unit. Several image objects in one analysis come from the same experimental plate. Treating each image as an independent experiment inflates the apparent sample size, whereas summarizing the images within each plate preserves the plate as the independent unit. Since the question does not identify what one row represents, the experimental unit remains unclear.
    \item NeuN effect sizes, ANOVA results, and power estimates are affected by whether measurements are independent or paired within the same animal. A NeuN power calculation treats left- and right-hemisphere measurements as independent observations, but the materials do not state whether the two measurements come from the same animal. If they are paired within animals, the analysis must account for that dependence, and the number of measurements cannot be interpreted as the number of independent animals. The reference calculation therefore makes an independence assumption without the subject-level pairing information needed to justify it.

\item Natural spline estimates can change with the observations included, the allowed flexibility of the curve, the model-selection rule, the prediction grid, and the method used to quantify uncertainty around the estimated maximum. In the swarming analysis, the frequency variable naturally includes the two pure-strain endpoints, but the reference notebook fits the curve after omitting them. The question does not instruct the analyst to exclude these observations, and the reference notebook does not explain the exclusion. Consequently, both the fitted curve and its estimated maximum depend on an unstated data-selection step.

\item Expression correlation analyses are influenced by score direction, expression transformation, low-expression filtering, sample matching, and the definition of gene length. In DepMap, a more negative GeneEffect value means stronger dependency. Some analyses reverse the score and use (-)GeneEffect, so that stronger dependency is represented by a larger value. Reversing the score also reverses the correlation: a positive correlation becomes negative, and a negative correlation becomes positive. Because the question does not specify whether to use GeneEffect or (-)GeneEffect, the direction of the correlation is not uniquely defined.

\item For miRNA analyses, important choices include qPCR quality control, the treatment of undetected Ct values, within-sample normalization, statistical testing, multiple-testing correction, and feature annotation. In one serum miRNA comparison, the source study normalizes each sample using the average Ct value of the miRNAs detected in that sample, whereas the reference notebook processes the Ct measurements differently before testing. These workflows analyze different quantities and can identify different sets of significant miRNAs. The normalization procedure is not specified in the question, and the reference notebook does not follow the processing method described in the source study.

\item ENO1 proteomics is comparatively stable because the requested values can be read from an existing results table, provided that the comparison direction, fold change, log2 fold change, and adjusted (p)-values are kept distinct. The ENO1 table contains a large tumor-to-normal effect estimate but a nonsignificant result after multiple-testing correction. This is not a hidden-parameter problem. The main risk is an interpretation error in which a large effect size is reported as statistically significant despite the adjusted result.

\item In m6A DEG analyses, the outcome depends on whether rows, peaks, transcripts, or unique genes are counted and on how repeated or conflicting records are combined. In the supplied m6A DEG table, one gene can appear in multiple peak or transcript rows, sometimes with conflicting methylation labels. The reference notebook reproduces row-level results and treats repeated records as separate observations, even when the question asks about genes. Without a gene-level aggregation rule, input rows may be treated as genes even though the two units are not equivalent.

\item CpG density calculations require genomic coordinates and chromosome lengths from the same species and genome assembly, followed by explicit rules for deduplication, density calculation, ratio direction, and chromosomes with zero counts. In the supplied files, the chromosome-length file labeled for Zebra Finch is identical to the Jackdaw file, and some reference outputs labeled as Zebra Finch are derived from Jackdaw quantities. An analysis can therefore run without error while combining one species' CpG coordinates with another species' chromosome information. This discrepancy originates from the packaged data and reference outputs rather than from a parameter choice made by the analyst.

\item Trimmomatic, BWA, and GATK outputs are shaped by complete command settings, software versions, reference files, coverage definitions, filtering status, and variant-counting units. In one trimming question, ``completely discarded'' can refer either to individual reads that are absent from all output files or to complete read pairs recorded as dropped in the processing log. These definitions count different units even when they refer to the same run. The intended unit is not defined, and no reference notebook or script is available to show how the original answer was obtained. The ambiguity therefore reflects both unclear wording and missing provenance.

\item Clustering and multiomics analyses are sensitive to duplicate handling, scaling, cluster number, random seed, label matching, and the definition of a complete sample. In one consensus-clustering workflow, the reference notebook does not set the random seed. Repeated runs can therefore use different random training and test splits and classify different samples as stable, even though the input data and stated settings remain unchanged. This unfixed stochastic step can directly change the result, while the question also does not require a seed or a stability rule that is robust across runs.

\item MAGeCK Reactome analyses are affected by how replicates are combined, how candidate genes and significance thresholds are defined, how gene identifiers are mapped, and which pathway resource and version are used. In one Reactome workflow, duplicated pathway definitions cause the same pathway to appear in multiple result rows, and the reference notebook counts those rows without making the duplication explicit. Counting unique pathway identifiers produces a different total. The difference reflects both the handling of duplicated pathway resources in the reference notebook and the absence of an explicit unique-pathway counting rule.

\end{itemize}

Specifically, we have compiled and summarized the questions corresponding to each error type in the dataset available at
\url{https://huggingface.co/datasets/ZhangxinMath/Bixbench-Verify}, where we also provide updated questions and their corresponding solutions. In many cases, the available information is not sufficient to establish a unique, reproducible result. This creates a broader evaluation problem: \textit{"a reported numerical result alone cannot show how the agent arrived at it, whether it followed the written instructions faithfully, or whether it used biological and statistical evidence to make a scientifically defensible choice that differed from the reference analysis."}

When an answer differs from the reference, the agent may simply have made a mistake. However, the disagreement may also arise because the agent believes it has made a more scientifically justified analytical choice than the one used in the reference analysis. Treating every disagreement as an error could therefore penalize a scientifically better analysis. Agreement with the reference is not conclusive either. Random variation, an unspecified seed, or version-specific software behavior may produce the same result without demonstrating reliable reasoning. A numerical match may also reproduce an analysis that is technically executable but scientifically weak.

A reference notebook is essential provenance because it records how a reference result was produced, but it should not automatically be treated as the only scientifically valid specification. A notebook may contain choices that are not required by the question, rely on mutable external resources, or conflict with the source study or packaged data. It can therefore document an implementation without fully justifying it.

Independent expert review is difficult because bioinformatics combines computational, biological, and statistical judgment. A computing expert may confirm that the code runs but may not be able to determine whether the outcome, experimental unit, or comparison is scientifically appropriate. A biological expert may identify a problem in the study design but may not be able to reproduce the computational environment. A bioinformatics expert can help connect these areas, but no single expert is likely to have deep knowledge of all relevant biological domains, statistical methods, and computational resources. Their assessment will still depend on their own training and experience. Complex tasks may therefore require the review of several specialists.

Because neither a reference notebook nor a single expert review is sufficient, future benchmarks should distinguish explicitly between \textbf{reproduction tasks} and \textbf{scientific-analysis tasks}. Reproduction tasks should require exact replication of a stated workflow and should pin the input files, software environment, database snapshots, and random seeds. Scientific-analysis tasks should define the biological target, analysis unit, minimum statistical requirements, and acceptable alternative specifications, while treating the notebook as one reference implementation rather than the sole source of truth. Evaluation should classify disagreements as agent errors, reasonable responses to underspecification, reference-implementation defects, data or provenance defects, or environment drift. It should score not only the final value but also the validity of the data selection, statistical model, assumptions, sensitivity analysis, and supporting computational evidence. Multidisciplinary review and adjudication should be used for cases that cannot be automatically resolved.

Prompt Chain would distinguish accidental numerical agreement from a reproducible and scientifically justified analysis, while avoiding the automatic rejection of an agent that reaches a different result for defensible reasons. The BixBench audit also directly informed the construction of our skills. Rather than hard-coding answers to erroneous or highly customized benchmark questions into the skills, we abstract recurring sources of ambiguity into reusable control points. For example, differential-expression skills explicitly track contrast direction, model formulas, low-expression filtering rules, and background gene sets; variant-analysis skills record denominator selection and normalization rules; and enrichment-analysis skills specify ranking criteria and universe definitions. When historical BixBench tasks genuinely require special conventions, we retain these as compatibility notes, while keeping the default behavior aligned with general-purpose bioinformatics practice. The resulting skills are therefore \textbf{benchmark-informed} rather than \textbf{benchmark-memorizing}.

To further validate the effectiveness of the revised evaluation set, we have re-tested the 67 modified questions using Bioinfoysis with a GPT-5.5 API. The system correctly answers 46 out of 67 questions, achieving a success rate of 68.66\%.

\subsection{Extended Evaluation on LAB-Bench 2}
\label{lab: Labbench2_extended_evaluation}

\textbf{Manual review scope.}
We have manually reviewed 91 cases from the SeqQA component of LAB-Bench2. Many of these cases are open-ended sequence-analysis and sequence-design tasks that permit multiple biologically valid answers. Therefore, they cannot be reliably evaluated through exact-match comparison against a single benchmark reference. For each case, we have assessed whether the response satisfies the biological and task-specific requirements.

\textbf{Evaluation criteria.}
The benchmark reference answer is treated as the grading target for deterministic tasks, including amino-acid lookup, mutation interpretation, and exact set or distance calculations. For sequence-design tasks, the reference answer is treated as one valid solution rather than the only permissible solution. An alternative primer, oligonucleotide, amplicon, or coding sequence is accepted if it satisfies the biological and task-specific constraints stated in the prompt. Depending on the task, these checks include sequence orientation, full coding-sequence coverage, translation into the requested protein, restriction-site compatibility, Gibson overlap validity, antisense complementarity, amplicon length or GC content, and primer-pair validity. A response is scored as incorrect if the reported sequence or value fails these checks, if a required answer is absent, or if the response contains only a placeholder or procedural instruction.

\textbf{Manual audit results.}
The audit has classified 63 of the 91 cases as correct and 28 as incorrect, corresponding to an accuracy of 69.23\% within the manually reviewed subset. No cases remain unresolved after review. Table~\ref{tab:seqqa-manual-audit} reports the outcomes by task type. The largest groups of accepted answers are valid codon-optimized coding sequences (18 cases), valid restriction-cloning primer pairs (14 cases), valid coding-sequence amplicons (12 cases), valid antisense oligonucleotides (9 cases), and valid alternative primer pairs for amplicon-GC tasks (6 cases). These results confirm that exact reference matching underestimates performance on tasks that admit multiple biologically valid solutions.

\begin{table}[t]
\centering
\caption{Manual review outcomes for the 91 SeqQA cases assigned \texttt{needs\_llm\_judge}. Accuracy is calculated within each task type.}
\label{tab:seqqa-manual-audit}
\small
\begin{tabular}{lrrrr}
\toprule
Task type & Cases & Correct & Incorrect & Accuracy (\%) \\
\midrule
Amplicon GC & 6 & 6 & 0 & 100.0 \\
Mutation restriction & 2 & 2 & 0 & 100.0 \\
Codon optimization & 19 & 18 & 1 & 94.7 \\
Amplicon length & 15 & 12 & 3 & 80.0 \\
Primer design & 19 & 14 & 5 & 73.7 \\
Oligo design & 13 & 9 & 4 & 69.2 \\
Primer interactions & 3 & 1 & 2 & 33.3 \\
Gibson primers & 7 & 1 & 6 & 14.3 \\
Synonymous mutation & 3 & 0 & 3 & 0.0 \\
ORF amino acid & 3 & 0 & 3 & 0.0 \\
Pairwise distances & 1 & 0 & 1 & 0.0 \\
\midrule
Total & 91 & 63 & 28 & 69.2 \\
\bottomrule
\end{tabular}
\end{table}

\textbf{Observed error modes.}
Among the 28 incorrect cases, the most frequent error is disagreement with a deterministic benchmark reference (7 cases). Other errors include the absence of a valid primer pair (5 cases), invalid antisense oligonucleotides (4 cases), invalid Gibson primer pairs (4 cases), amplicons that do not fully cover the requested coding sequence (3 cases), placeholder or instruction-only responses (2 cases), incorrect sets (2 cases), and an invalid or missing codon-optimized sequence (1 case). These failure modes distinguish semantic answer errors from answer-extraction failures.



\begin{table}[t]
    \centering
    \caption{Performance of Bioinfoysis on the SeqQA2 evaluation across four model families, reported under the revised grading protocol. Agent accuracy denotes the full Bioinfoysis workflow; base-model accuracy denotes the Direct condition (API with attachment injection / file tools).}
    \label{tab:seqqa2-accuracy}
    \resizebox{\linewidth}{!}{%
    \begin{tabular}{l r r r}
        \toprule
        Model & Agent accuracy (\%) & Base-model accuracy (\%) & Difference (pp) \\
        \midrule
        DeepSeek-V4 & 65.25 (261/400) & 25.25 (101/400) & +40.00 \\
        GLM-5.2     & 67.50 (270/400) & 35.00 (140/400) & +32.50 \\
        GPT-5.5     & 71.25 (285/400) & 63.50 (254/400) & +7.75 \\
        Kimi-2.6    & 62.50 (250/400) & 25.50 (102/400) & +37.00 \\
        \bottomrule
    \end{tabular}%
    }
\end{table}

\textbf{Interpretation.}
Under the revised grading protocol, the SeqQA2 results preserve a consistent advantage of Bioinfoysis over the corresponding base-model-only (Direct) condition across all four model families. The Agent configurations achieved accuracies from 62.50\% to 71.25\%, compared with 25.25\% to 63.50\% for the base models. The largest improvement occurred for DeepSeek-V4 (+40.00 percentage points), followed by Kimi-2.6 (+37.00) and GLM-5.2 (+32.50), while GPT-5.5 showed the smallest but still substantial improvement (+7.75). GPT-5.5 ranks first and GLM-5.2 second under both conditions (Agent: 71.25\% and 67.50\%; Direct: 63.50\% and 35.00\%), while DeepSeek-V4 and Kimi-2.6 remain at the bottom in the Direct setting, separated by only 0.25 percentage points (25.25\% vs.\ 25.50\%). The three non-GPT models cluster closely together in the Agent setting (62.50\%--67.50\%) but substantially lower in the Direct setting (25.25\%--35.00\%). This pattern indicates that the harness contributes most where base-model execution and answer-formatting reliability are weakest; the comparatively small gain for GPT-5.5 suggests that its stronger intrinsic instruction following and output-format compliance already capture much of what the workflow provides. Because both conditions were re-scored on the same 400-question set with the same revised grading protocol, the observed differences reflect paired evaluation conditions rather than differences in task coverage or grading criteria.



\end{document}